\documentclass[10pt,letterpaper]{article}
\usepackage[top=0.85in,left=2.75in,footskip=0.75in]{geometry}
\usepackage{booktabs,makecell,adjustbox}

\usepackage{amsmath,amssymb,amsfonts}

\usepackage{changepage}

\usepackage{textcomp,marvosym}
\usepackage{enumitem}

\usepackage{cite}

\usepackage{nameref,hyperref}
\usepackage{lastpage}

\usepackage[right]{lineno}

\usepackage[nopatch=eqnum]{microtype}
\DisableLigatures[f]{encoding = *, family = * }

\usepackage[table]{xcolor}

\usepackage{array}

\usepackage{algorithm}
\usepackage{algpseudocode}

\usepackage{graphicx}
\usepackage{float}
 \usepackage{comment}
\usepackage{multirow}
\usepackage{booktabs}
\usepackage{pgf-pie}
\usepackage{tikz}

\newcolumntype{+}{!{\vrule width 2pt}}
\newlength\savedwidth

\raggedright
\usepackage[aboveskip=1pt,labelfont=bf,labelsep=period,justification=raggedright,singlelinecheck=off]{caption}

\makeatletter
\renewcommand{\@biblabel}[1]{\quad#1.}
\makeatother

\usepackage{lastpage,fancyhdr,graphicx}
\usepackage{epstopdf}
\fancyheadoffset[L]{2.25in}
\fancyfootoffset[L]{2.25in}
\begin{document}

\begin{flushleft}
{\Large
\textbf{Learning to Undo: Rollback-Augmented Reinforcement Learning with Reversibility Signals} 
}
\newline
\\
Andrejs Sorstkins\textsuperscript{1\Yinyang*},
Omer Tariq\textsuperscript{2\Yinyang*},
Muhammad Bilal\textsuperscript{1\ddag*},

\bigskip
\textbf{1} School of Computing and Communications, Lancaster University, Lancaster LA1~4WA, United Kingdom
\\
\textbf{2} Neubility, 2F 115 (04768) Wangsimni-ro, Seongdong-gu, Seoul, South Korea
\\
\bigskip

%
%
\Yinyang These authors contributed equally to this work.\\
\ddag Supervisor, editor and steering professor \\

*info@benarktech.co.uk; omertariq@kaist.ac.kr; m.bilal8@lancaster.ac.uk;

\end{flushleft}


%

\section*{Abstract}
This paper postulates a novel reversible learning framework designed to enhance the robustness and efficiency of value-based Reinforcement Learning (RL) agents, specifically addressing their pervasive vulnerability to value overestimation and instability in partially irreversible environments. The framework instantiates two complementary core mechanisms: an empirically derived transition reversibility measure ($\Phi$(s, a)) and a selective state-rollback operation.
To achieve this, we introduce an online, per-state-action estimator ($\Phi$) that quantifies the likelihood of returning to a prior state within a fixed horizon K. This measure is used to adjust the penalty term during temporal difference updates dynamically, integrating reversibility awareness directly into the value function. Crucially, the system incorporates a selective rollback operator: when an action yields an expected return markedly lower than its instantaneous estimated value (violating a predefined threshold), the agent is penalized and reverts to the preceding state rather than progressing. This strategically interrupts sub-optimal, high-risk trajectories and avoids catastrophic steps.
By synergistically combining this reversibility-aware evaluation with targeted rollback, the proposed methodology demonstrably improves safety, performance, and stability. Empirically, in the CliffWalking-v0 domain, the framework reduced catastrophic falls by over 99.8\% and yielded a 55\% increase in mean episode return. Similarly, in the Taxi-v3 domain, it suppressed illegal actions by $\geq99.9\%$ and achieved a 65.7\% improvement in cumulative reward, while also sharply reducing reward variance in both environments. Ablation studies confirm the rollback mechanism is the critical component underlying these substantial safety and performance gains, marking a robust step toward safe and reliable sequential decision-making.


\section{Introduction}

Reinforcement learning (RL) paradigms have demonstrated state-of-the-art efficacy across an array of domains, from strategic board games such as Go and discrete control benchmarks like Atari, to real-world control problems in high-dimensional robotics and complex, unstructured environments \cite{sutton2018reinforcement}. The remarkable performance gains achieved by deep RL methods-through innovations in function approximation, experience replay, and actor–critic architectures-have reignited interest in deploying such algorithms for real-world decision-making tasks. Nevertheless, the transition of RL algorithms from controlled experimental settings to operational environments is frequently impeded by training-induced instability, sample inefficiency, and emergent unsafe behaviors \cite{amodei2016concrete}. A primary factor contributing to these challenges is the pervasive overestimation of action-value functions \cite{thrun1993issues}, which skews policy improvement towards trajectories with spuriously optimistic reward predictions. When an agent’s \(Q\)-function becomes biased towards overly optimistic reward estimates \cite{vanhasselt2015double}, it preferentially pursues statistically spurious or low-probability trajectories, precipitating oscillatory policy updates, prolonged convergence times, and, in worst-case scenarios, catastrophic failures within safety-critical infrastructures such as aerospace control systems or nuclear facility management.

Reversibility, an intrinsic aspect of human cognitive architectures, underpins our capacity for deliberative decision-making and adaptive learning. Individuals habitually assess not only the immediate reward associated with a given action, but also the extent to which that action can be reversed or counteracted by subsequent steps. This involves generating internal counterfactual simulations-mental “rollbacks”-to evaluate potential failure modes and to hedge against irreversible outcomes. Such meta-cognitive processes enable humans to engage in risk-sensitive exploration, to remediate mistakes via corrective maneuvers, and to maintain a consistent trajectory towards long-term objectives. Despite its foundational relevance to algorithmic safety and robustness, this latent human impulse to “undo” suboptimal decisions-and thereby explore alternative strategies without irrevocable consequence-remains scarcely addressed in existing RL research frameworks.

Embedding reversibility into an RL framework offers an illustrative principle for a broad spectrum of safety-critical applications \cite{garcia2015comprehensive,GuerrierFouadBeltrame2024_CBF_survey,ZhaoHeChenWeiLiu2023_statewise_survey,WachiShenSui2024_constraint_survey,Kushwaha_Ravish_Lamba_Kumar_2025,gu2024_review_safeRL}. Consider autonomous vehicular control, where irreversible errors-such as collisions-can precipitate loss of life or property damage; or robotic surgical assistants, where miscalibrated manipulations must be promptly retracted to avoid patient harm. Similarly, adaptive medical treatment planning algorithms must be able to backtrack from harmful dosage adjustments, and industrial process control systems must swiftly revert hazardous state transitions to prevent environmental or infrastructural compromise. In these contexts, the inability to retract or attenuate deleterious transitions can incur unacceptable risk. We address this exigency by integrating an online reversibility estimator-a learned function that predicts the probability of returning to a safe state distribution from any given transition-with an explicit rollback operator. Upon detection of high-risk transitions-quantified via this reversibility metric-the system effectuates a corrective “U-turn,” restoring the agent to a prior checkpointed state. This mechanism not only constrains exploratory risk and prevents agent entrapment in irreversible error states but also attenuates policy divergence, thus facilitating stable convergence under rigorous safety constraints.

Conventional cures for \(Q\)-overestimation-dual/twin critics, bias-corrected evaluation, and conservative \(Q\)-learning-often trade accuracy for added critics, tighter update rules, and cautious behavior, inflating compute and sample cost \cite{lan2020maxmin}. In reversibility-aware RL, \cite{grinsztajn2021reversibility} learn a “precedence” score from raw trajectories to avoid irreversible regions, but their approach trains a Siamese classifier tied to the behavior policy, uses a fixed temporal window, relies on a global threshold to gate actions, and never actually undoes a damaging step. We address these gaps with a rollback-augmented framework that couples (i) a per-state–action empirical reversibility estimator \(\Phi(s,a)\), computed online via a FIFO return-within-\(K\) test and updated by a light EMA, with (ii) an explicit “U-turn” rollback that fires only when the reversibility-penalized TD target falls below a threshold; \(\Phi\) also induces a localized penalty in the TD update. This design eliminates the need for a learned Siamese model, adapts naturally to different horizons via \(K\), replaces a blunt global irreversibility proxy with per-state–action estimates, and, crucially, equips the agent with an actionable undo. Empirically, in \texttt{CliffWalking-v0} we cut catastrophic falls by \(\geq 99.8\%\) and improve mean return by \(\sim 55\%\) while collapsing return variance by \(\sim 71\%\); in \texttt{Taxi-v3} we suppress illegal actions by \(\geq 99.9\%\) with \(\sim 66\%\) return gains and \(\sim 59\%\) variance reduction. Our contributions are: (1) a scalable, model-free, per-state–action reversibility estimator that avoids classifier training; (2) an explicit rollback operator integrated into tabular \(Q\)-learning and SARSA updates; (3) a principled coupling of \(\Phi\)-shaping and selective rollback that bounds downside without choking exploration; and (4) extensive evaluation, sensitivity analyses, and ablations that isolate which components matter for safety and performance.

\section{Background}

Reinforcement learning has made incremental improvements over the last few decades. Overestimation of action values has long been recognized as a key obstacle to stable and efficient learning in value-based RL, making it one of the main challenges to address. In this section, we provide a brief overview of the key foundational concepts.

\subsection{Reinforcement Learning and Markov Decision Processes}

Reinforcement learning (RL) frames sequential decision-making as a Markov decision process (MDP) \cite{sutton2018reinforcement}
\begin{align}
  (\mathcal{S}, \mathcal{A}, P, R, \gamma),
\end{align}
where an agent in state \(s \in \mathcal{S}\) chooses action \(a \in \mathcal{A}\), receives reward \(r\), and transitions to \(s' \sim P(\cdot \mid s,a)\) with discounted return
\begin{align}
  G_t = \sum_{k=0}^{\infty} \gamma^k r_{t+k}.
\end{align}
Value-based RL approximates the action-value function
\begin{align}
  Q(s,a) \approx \mathbb{E}[\,G_t \mid s_t = s, a_t = a\,]
\end{align}
via temporal-difference updates.

\subsection{Tabular Q-Learning}

\(Q\)-learning \cite{watkins1989learning} updates a table of values via
\begin{align}
  Q(s,a) \leftarrow Q(s,a) + \alpha \bigl(r + \gamma \max_{a'} Q(s',a') - Q(s,a)\bigr).
\end{align}
This off-policy rule can converge to the optimal action-value function under sufficient exploration.

\subsection{SARSA}

SARSA is on-policy: it updates toward the value of the action actually taken, sampling \(a' \sim \pi(\cdot \mid s')\):
\begin{align}
  Q(s,a) \leftarrow Q(s,a) + \alpha \bigl(r + \gamma Q(s',a') - Q(s,a)\bigr).
\end{align}
This ensures updates remain consistent with the agent’s current policy.

\subsection{Selection and Evaluation}

Early work tackled this by decoupling selection and evaluation: Double \(Q\)-Learning \cite{vanhasselt2010doubleq} maintains two independent estimators-using one to choose actions and the other to evaluate them-curbing maximization bias in both tabular and deep settings \cite{vanhasselt2015double}. TD3 extends this idea to continuous control by clipping between twin critics and delaying policy updates to further suppress overoptimism \cite{fujimoto2018addressing}. Rather than relying solely on multiple networks, Maxmin \(Q\)-Learning maintains \(N\) critics and interpolates between their highest and lowest predictions via a tunable parameter \(\kappa\). By adjusting \(\kappa\), it trades off optimism against conservatism, yielding tighter theoretical bounds on estimation error and empirical gains across benchmarks \cite{lan2020maxmin}.

\subsection{Precedence Estimation}

As introduced by Grinsztajn N.\cite{grinsztajn2021reversibility}, \emph{precedence} is a self-supervised statistic capturing the temporal “direction” between two states under a fixed policy \(\pi\) and horizon \(T\). They define
\begin{align}
  \psi_{\pi,T}(s,s')
  = \frac{\mathbb{E}_{\tau \sim \pi}\bigl[\,\lvert\{(t,t') : t'<t,\; s_t = s,\; s_{t'} = s'\}\rvert\bigr]}
         {\mathbb{E}_{\tau \sim \pi}\bigl[\,\lvert\{(t,t'): t \neq t',\; s_t = s,\; s_{t'} = s'\}\rvert\bigr]},
\end{align}
estimating the probability that state \(s\) appears after \(s'\) in trajectories of length \(\le T\). In practice, one samples trajectories, collects state-pairs within a window \(\lvert t - t' \rvert \le w\), and computes the fraction with \(t' < t\):
\begin{itemize}
  \item If \(\psi \approx 1\), transitions \(s \to s'\) are essentially irreversible.
  \item If \(\psi \approx 0.5\), no consistent ordering exists, indicating reversibility.
\end{itemize}

They then lift \(\psi\) to an action-level score by averaging over next-state distributions:
\begin{align}
  \bar\phi_\pi(s,a) = \mathbb{E}_{s' \sim P(\cdot \mid s,a)}\bigl[\psi_{\pi,T}(s',s)\bigr],
\end{align}
which serves as a data-driven proxy for reversibility without external labels or models.

\subsection{Related Work and Comparative Positioning}

Although these approaches each mitigate overestimation in different ways-through alternate estimators, bias–variance blending, or learned state–action reversibility they stop short of explicitly undoing poor decisions. Safe exploration approaches in MDPs \cite{moldovan2012safe} similarly aim to avoid irreversible failures, but do not provide rollback mechanisms.

Safe exploration in RL has been widely studied due to the risks of unsafe behavior during training and deployment. Existing approaches can be grouped into three broad categories: (i) \emph{constraint-based formulations}, (ii) \emph{verification-based methods}, and (iii) \emph{optimization-based trade-off techniques}.

\subsubsection{Constraint-based safe exploration}
Wachi et al.~\cite{wachi2023gse} introduce the \emph{Generalized Safe Exploration (GSE)} framework, which unifies common safe RL formulations-cumulative, state, and instantaneous constraints-into a meta-algorithm (MASE) with high-probability safety guarantees. By penalizing unsafe actions before actual violations, MASE ensures safety even during training, extending beyond average-case constraint satisfaction. 
Similarly, As et al.~\cite{as2025actsafe} propose \emph{ActSafe}, a model-based approach that learns probabilistic dynamics models and couples optimistic exploration with pessimistic safety constraints. ActSafe provides finite-sample complexity guarantees while scaling to high-dimensional deep RL settings.

\subsubsection{Verification-based safe RL.}
Formal verification methods have also been applied to ensure provable safety during exploration. Wang and Zhu~\cite{wang2024velm} propose \emph{VELM} (Verified Exploration through Learned Models), which learns symbolic environment models amenable to reachability analysis. VELM constructs a shielding mechanism that confines the agent’s actions to formally verified safe regions, thereby reducing violations without degrading reward performance. While powerful, such approaches often depend on the tractability of symbolic regression or approximations of nonlinear dynamics, which can limit applicability in highly stochastic or large-scale domains.

\subsubsection{Reward–safety trade-off optimization.}
Another line of research emphasizes balancing safety constraints with performance. Gu et al.~\cite{gu2025gradient} highlight the intrinsic gradient conflict between reward maximization and safety optimization. Their framework introduces gradient manipulation techniques to reconcile these conflicts, producing improved trade-offs across Safety-MuJoCo and OmniSafe benchmarks. This direction complements earlier constrained optimization methods (e.g., CPO \cite{achiam2017cpo}, PPO-Lagrangian \cite{ray2019benchmarking}), but with a sharper focus on handling conflicting optimization signals.

\subsubsection{Positioning of this work.}
In contrast to prior approaches that either enforce hard constraints (e.g., GSE, VELM) or resolve gradient conflicts \cite{gu2025gradient}, our work introduces a \emph{reversibility-driven perspective}. We propose an empirical reversibility estimator coupled with a rollback operator that enables the agent not only to avoid unsafe regions but to actively \emph{undo} detrimental steps.(Algorithm ~\ref{alg:multi_knapsack}).This mechanism provides an additional layer of resilience absent in most existing safe exploration frameworks, which typically rely on forward-looking predictions or static safety filters. Unlike ActSafe, which guarantees safety by conservative set expansion, our rollback mechanism offers \emph{dynamic recoverability}, making exploration less brittle in environments where occasional missteps are unavoidable. Moreover, by empirically demonstrating over 99\% reduction in catastrophic actions and consistent return improvements, our method complements existing safe RL approaches by offering a pragmatic, model-free safeguard against irreversible outcomes.

\begin{algorithm}[H]
\caption{Modified Q-Learning with Precedence and rollback}\label{alg:multi_knapsack}
\begin{algorithmic}[1]
  \Require $Q[s,a]\gets Q_{0},\;\Phi[s,a]\gets \phi_{0},\;\text{buffer}\gets \emptyset,\;t\gets 0$
  \While{true}
    \State $a \gets \epsilon$\texttt{-greedy}$(Q[s,\cdot])$
    \State observe reward $r$, next state $s'$, and flag $\mathit{done}$
    \State $t \gets t + 1$
    \medskip
    \ForAll{records $(s_{0},a_{0},d)$ in buffer}
      \If{$s' = s_{0}$}
        \State $y \gets 1$
      \ElsIf{$t > d$}
        \State $y \gets 0$
      \Else
        \State \textbf{continue}
      \EndIf
      \State $\Phi[s_{0},a_{0}] \gets (1-\alpha_{\phi})\,\Phi[s_{0},a_{0}] + \alpha_{\phi}\,y$
      \State remove $(s_{0},a_{0},d)$ from buffer
    \EndFor
    \State append $(s, a, t + K)$ to buffer
    \State $r' \gets r - \lambda\,\bigl(1 - \Phi[s,a]\bigr)$
    \If{$\mathit{done}$}
      \State $\mathit{target} \gets r'$
    \Else
      \State $\mathit{target} \gets r' + \gamma\,\max_{a'}Q[s',a']$
    \EndIf
    \State $\delta \gets \mathit{target} - Q[s,a]$
    \If{$\mathit{target} \le T \cdot Q[s,a]$}
      \State $\beta' \gets \beta,\;\text{rollback}\gets\text{true}$
    \Else
      \State $\beta' \gets 1,\;\text{rollback}\gets\text{false}$
    \EndIf
    \State $Q[s,a] \gets Q[s,a] + \alpha\,\beta'\,\delta$
    \If{rollback \textbf{and} $\neg\,\mathit{done}$}
      \State $s \gets s$ 
    \Else
      \State $s \gets s'$
    \EndIf
  \EndWhile
\end{algorithmic}
\end{algorithm}

\begin{figure}[H]
  \centering
  \includegraphics[width=\columnwidth,trim=40 60 40 60,clip]{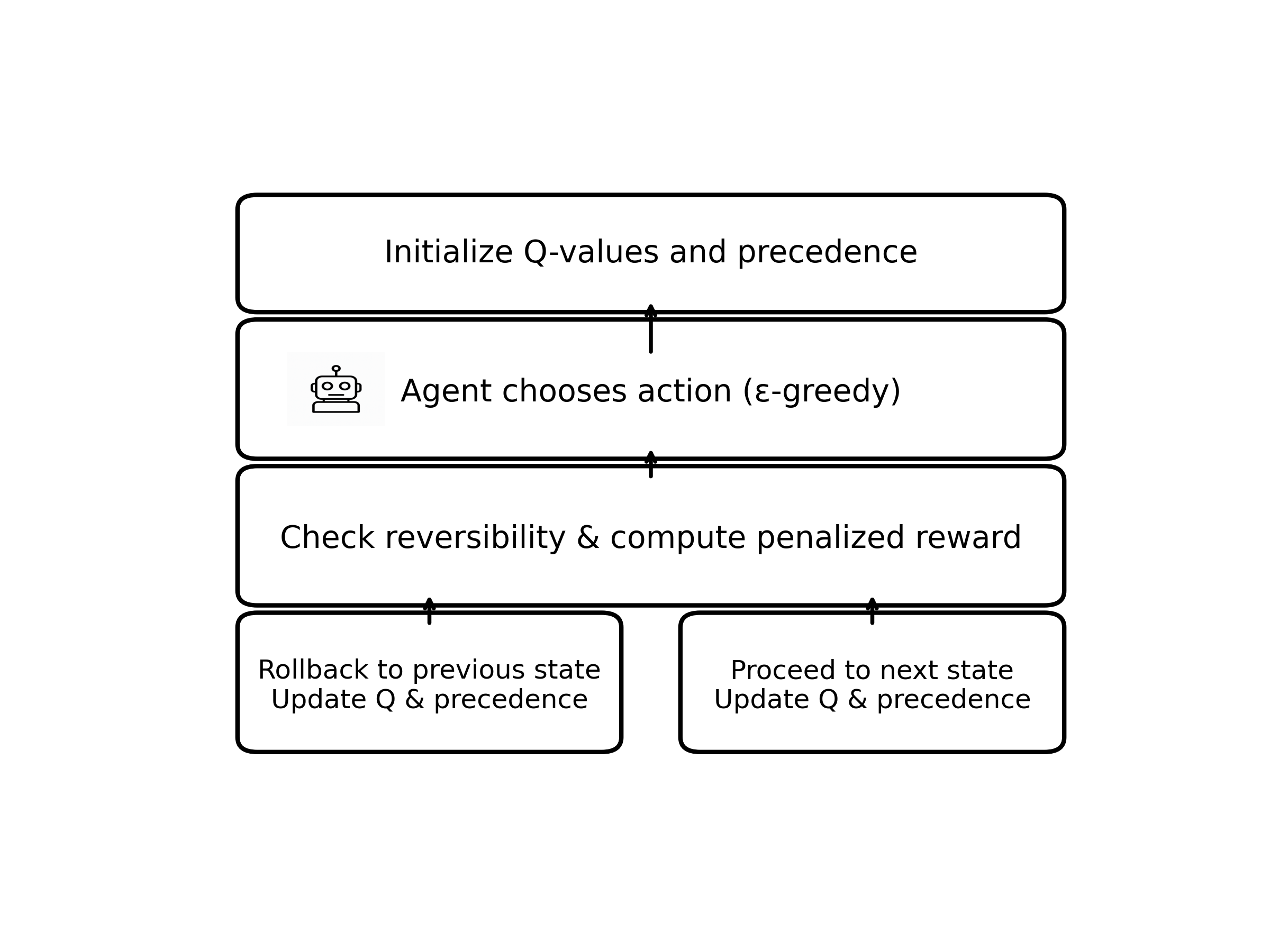}
  \caption{Reversible \(Q\)-learning with rollback and precedence.}
  \label{fig:revrl}
\end{figure}

\section{System Overview}
\label{sec:system-overview}

\noindent
Precedence-based reversibility \cite{grinsztajn2021reversibility} offers a self-supervised signal for whether transitions can “undo” themselves, yet it exhibits four coupled weaknesses in practice. First, the Siamese classifier is trained purely on the agent's own trajectories and can overfit to policy-specific quirks; if the behavior policy is near-deterministic or fails to revisit certain state–action pairs, the estimator may systematically mislabel reversible transitions as irreversible (and vice versa). Second, a fixed temporal window \(w\) forces short-horizon judgments and obscures longer-range reversibility that requires extended return paths. Third, both the Reversibility-Aware Explorer (RAE) and Controller (RAC) rely on a single global threshold \(\beta\) to gate actions, a blunt control that struggles in heterogeneous state–action spaces and requires environment-specific retuning. Fourth, neither RAE nor RAC provides an explicit \emph{rollback} mechanism; once a damaging move is taken, the agent cannot immediately undo it, leaving learning exposed when some irreversible steps are unavoidable. Related work on skill discovery has also implicitly leveraged reversibility, for example in unsupervised RL approaches such as DIAYN \cite{eysenbach2019diversity}, where diversity-enforcing objectives yield reusable and often reversible behaviors. However, these works do not provide explicit rollback mechanisms.

\paragraph{Our approach.}
We replace classifier-based precedence with a lightweight, empirical reversibility estimate maintained online and coupled to an explicit rollback operator (Fig~\ref{fig:revrl}). Rather than fitting a Siamese model to policy-induced data, we enqueue each observed transition into a fixed-size FIFO structure of length \(K\) and update a per-state–action estimate \(\Phi(s,a)\) via an exponential moving average. The horizon is thus controlled solely by \(K\), allowing short- or long-range reversibility without retraining. We integrate \(\Phi\) into temporal-difference (TD) learning through a localized penalty that is applied only when a \emph{reversibility-penalized} target breaches a threshold. Because \(\Phi\) is defined per \((s,a)\), this yields fine-grained, data-driven shaping rather than a single global cutoff. Crucially, when the same threshold condition is violated, we execute an explicit rollback that resets the agent to the previous state, preventing irreversible missteps from contaminating subsequent learning. This corrective principle is reminiscent of off-policy correction methods such as \(Q(\lambda)\) with importance adjustments \cite{harutyunyan2016qlambda}, but in contrast, our rollback mechanism directly intervenes at the state-transition level rather than adjusting the weighting of returns.

\subsection{Empirical Reversibility via a Precedence FIFO}
\label{subsec:fifo}

Consider a transition \((s_t, a_t)\rightarrow s_{t+1}\). Immediately after observation, we push a \emph{pending record} \((s_0, a_0, d)\) onto a FIFO list \(L\), where \(s_0=s_t\), \(a_0=a_t\), and \(d=t+K\) is a deadline by which a return to \(s_0\) must occur to be counted as reversible. On each subsequent step, we scan pending records in \(L\):
(i) if the current state matches any \(s_0\) before its deadline, we set \(y=1\) and remove that record; (ii) if the deadline is exceeded without a match, we set \(y=0\) and remove it; (iii) otherwise, the record remains pending. Because each record is dequeued no later than \(K\) steps after insertion, \(|L|\le K\) and memory is bounded.

When a record resolves with label \(y\in\{0,1\}\), we update the reversibility table \(\Phi:\mathcal{S}\times\mathcal{A}\to[0,1]\) by an exponential moving average (EMA):
\begin{equation}
\label{eq:phi-update}
\Phi[s_0,a_0] \leftarrow (1-\alpha_\phi)\,\Phi[s_0,a_0] + \alpha_\phi\, y,
\end{equation}
with small learning rate \(\alpha_\phi\ll 1\). Under stationarity and sufficient visitation, the EMA converges to the probability of returning to \(s_0\) within \(K\) steps. Intuitively, frequent returns drive \(\Phi\to 1\) (high reversibility), whereas persistent non-returns drive \(\Phi\to 0\) (high irreversibility). Initialization of \(\Phi\) encodes prior risk posture: pessimistic (\(\Phi\approx 0\)), neutral (\(\Phi\approx 0.5\)), or optimistic (\(\Phi\approx 1\)); we study these priors empirically to simulate different exploration biases.

\subsection{TD Learning with Penalization and Rollback}
\label{subsec:td-rollback}

We maintain two tabular objects: the action-value function \(Q[s,a]\) and the reversibility estimate \(\Phi[s,a]\). At each step, we form a \emph{penalized reward}
\begin{equation}
\label{eq:penalized-reward}
r' \;=\; r \;-\; \lambda\,\bigl(1-\Phi[s_t,a_t]\bigr),
\end{equation}
where \(\lambda\ge 0\) scales the irreversibility penalty. This yields the modified TD error for \(Q\)-learning
\begin{equation}
\label{eq:td-q}
\delta \;=\; r' + \gamma \max_{a'} Q(s_{t+1},a') - Q(s_t,a_t),
\end{equation}
and for SARSA
\begin{equation}
\label{eq:td-sarsa}
\delta \;=\; r' + \gamma Q(s_{t+1},a_{t+1}) - Q(s_t,a_t).
\end{equation}

We introduce a multiplicative factor \(\beta\) to amplify corrections when the (unpenalized) target underperforms the current estimate by more than a threshold \(T\in(0,\infty]\):
\begin{equation}
\label{eq:beta}
\beta \;=\;
\begin{cases}
P, & \text{if } r + \gamma \max_{a'} Q(s_{t+1},a') \le T\,Q(s_t,a_t) \quad \text{(Q-learning)},\\[3pt]
P, & \text{if } r + \gamma Q(s_{t+1},a_{t+1}) \le T\,Q(s_t,a_t) \quad \text{(SARSA)},\\[3pt]
1, & \text{otherwise,}
\end{cases}
\end{equation}
with \(P\in(0, \infty]\) the penalty level used only in adverse targets. The value update is then
\begin{equation}
\label{eq:q-update}
Q(s_t,a_t) \leftarrow Q(s_t,a_t) + \alpha\,\beta\,\delta.
\end{equation}

\paragraph{Rollback operator.}
When the threshold condition in Eq~\eqref{eq:beta} is triggered, we execute a rollback by setting the next state to the current state (and, for SARSA, the next action to the current action).  
For \(Q\)-learning, the rollback operator is defined as
\begin{equation}
\label{eq:rollback-q}
s_{\text{next}} =
\begin{cases}
s_t,     & \text{if the threshold is violated}, \\
s_{t+1}, & \text{otherwise.}
\end{cases}
\end{equation}
For SARSA, the operator extends to both state and action:
\begin{equation}
\label{eq:rollback-sarsa}
(s_{\text{next}},a_{\text{next}}) =
\begin{cases}
(s_t,a_t),         & \text{if the threshold is violated}, \\
(s_{t+1},a_{t+1}), & \text{otherwise.}
\end{cases}
\end{equation}
This explicit “U-turn” prevents low-quality, potentially irreversible transitions from propagating errors and stabilizes exploration under risk.

\subsection{Design Rationale and Behavioral Control}
\label{subsec:rationale}

The FIFO construction bounds memory and enforces a clear \(K\)-step notion of reversibility; increasing \(K\) models higher “patience” before declaring a transition irreversible. The per-state–action \(\Phi\) produces localized penalties via Eq~\eqref{eq:penalized-reward}, in contrast to a global \(\beta\) cutoff in prior work; this improves compatibility with heterogeneous state–action topologies. The thresholded scaling \(\beta\) in Eq~\eqref{eq:beta} sharpens corrective updates only when warranted, avoiding chronic pessimism. Finally, the rollback in Eq~\eqref{eq:rollback-q} and Eq~\eqref{eq:rollback-sarsa} adds an \emph{actionable} recovery primitive absent from precedence-only schemes, reducing contamination from catastrophic steps. Together, these components yield a single, continuous process that neutralizes the four weaknesses of precedence-based reversibility without heavyweight classifiers or environment models.

\subsection{Interpreting Hyperparameters}
\label{subsec:hyperparams}

The horizon \(K\) governs the reversibility granularity and indirectly the agent's tolerance for delayed recovery; smaller \(K\) yields conservative, short-horizon judgments, while larger \(K\) captures longer detours. The initialization of \(\Phi\) encodes prior risk appetite (pessimistic, neutral, optimistic) and can be selected to match domain priors or safety requirements. The threshold \(T\) controls the acceptance level before rollback: higher \(T\) triggers rollbacks sooner (safer but potentially slower learning), while lower \(T\) tolerates temporary degradation to preserve exploration. We study sensitivity to \((K,\lambda,T,P,\alpha_\phi)\) in Section~\ref{sec:experiments}.

\section{Simulation}

\subsection{Environments}

All experiments were conducted using Gymnasium v1.2.0 (Farama Foundation, 2025)\footnote{\url{https://gymnasium.farama.org}}. This framework extends the original OpenAI Gym API \cite{brockman2016openai}, which remains a standard benchmark suite for reproducible reinforcement learning research. While our study focuses on single-agent tabular domains, similar reproducibility concerns have motivated the development of multi-agent environments such as PettingZoo \cite{terry2020pettingzoo}, which extends the Gym interface to multi-agent RL. Two canonical tabular “toy-text” domains were chosen to evaluate the reversible-RL algorithm under diverse yet tractable conditions:

\begin{enumerate}[label=\arabic*.]
  \item \textbf{Cliff Walking (\texttt{CliffWalking-v0}):}  
    A deterministic \(4\times 12\) grid with start at \([3,0]\) and goal at \([3,11]\). A “cliff” spans \([3,1]\)–\([3,10]\); stepping into it yields \(-100\) and teleports the agent back to start. Each regular step yields \(-1\), and the episode terminates upon reaching the goal. The observation space has 48 reachable states, and the action space has 4 discrete moves.

  \item \textbf{Taxi (\texttt{Taxi-v3}):}  
    A \(5\times 5\) grid in which a taxi must pick up a passenger at one of four fixed locations and deliver them to a specified destination. The observation space has size \(|\mathcal{S}|=500\) (25 taxi positions \(\times\) 5 passenger locations \(\times\) 4 destinations), and the action space \(|\mathcal{A}|=6\) (move south, north, east, west; pick up; drop off). Each step yields \(-1\); illegal pick-up/drop-off yields \(-10\); successful drop-off yields \(+20\). Episodes end upon successful passenger delivery.
\end{enumerate}

\subsection{Implementation Details}

All algorithms were implemented in Python~3.9 with Gymnasium~1.2.0 and NumPy~1.23, ensuring a pure tabular setting.

\subsection{Experimental Protocol}

All experiments employed a training budget of 100\,000 independent episodes per environment. Each episode in Cliff Walking and Taxi was executed until the agent reached the goal state or a 700-step time limit was reached in the Cliff Walking environment and a 1500-step limit in the Taxi environment, such that the cumulative negative rewards model “suffering” that the agent minimizes. Rollback counts as a step even when no state change occurs. A fixed sequence of random seeds was applied systematically across all episodes and agents to ensure each algorithm experienced identical stochastic conditions. Statistical information-including episodic returns, rollback counts, and convergence metrics-was recorded for all 100\,000 episodes and aggregated into CSV files for comprehensive post-hoc analysis. 

\subsection{Scope Justification}

Many recent advances in reinforcement learning target high-dimensional or continuous control tasks; our study deliberately focuses on tabular environments to rigorously evaluate the proposed reversibility framework. Tabular benchmarks such as \texttt{CliffWalking-v0} and \texttt{Taxi-v3} allow us to isolate the effects of our empirical reversibility estimator and U-turn rollback mechanism without the confounding complexities introduced by function approximation or representation learning. By removing factors like neural network training dynamics and policy-gradient variance, we can precisely quantify how reversibility influences both safety (e.g., reduction in catastrophic transitions) and performance (e.g., steady improvement in cumulative return). Moreover, the deterministic nature of tabular implementations ensures complete reproducibility: every buffer update, estimator statistic, and rollback decision can be logged and inspected in full. 

\section{Experimental Evaluation and Results}
\label{sec:experiments}

In this subsection, we evaluate the impact of integrating reversibility and rollback into the \(Q\)-learning framework, focusing on mean performance, safety outcomes, and variance control in both \texttt{CliffWalking-v0} and \texttt{Taxi-v3} environments (Table~\ref{tab:q_mod_table}). All reported statistics are computed over 100\,000 episodes per agent configuration; 95\% confidence intervals are reported as \(\bar x \pm 1.96\,\sigma/\sqrt{N}\). Rollback counts as a step for reward accounting in the next sub-sections.

\begin{table}[!ht]
\normalsize
\setlength{\tabcolsep}{3pt}
\renewcommand{\arraystretch}{1.15}

\begin{adjustwidth}{-2.0in}{-0.25in}

  \setbox0=\hbox{%
    \begin{adjustbox}{max width=\linewidth} 
      \begin{tabular}{@{}llrrrrrrrr@{}}
        \toprule
        \textbf{Domain} & \textbf{Metric} &
        \makecell{\textbf{Vanilla Q}\\\textbf{mean (CI)}} & $\sigma_{\mathrm{van}}$ &
        \makecell{\textbf{Rollback Precedence}\\\textbf{Q mean (CI)}} & $\sigma_{\mathrm{mod}}$ &
        \textbf{$\Delta$ mean} & \textbf{\%$\Delta$ mean} & \textbf{$\Delta\sigma$} & \textbf{\%$\Delta\sigma$} \\
        \midrule
        \multirow{4}{*}{\textbf{CliffWalking-v0}}
          & Total Reward           & $-399.77\,[-403.26,-396.27]$ & 563.78 & $-179.81\,[-180.81,-178.81]$ & 160.97 & $+219.96$   & $+55.0\%$ & $-402.81$ & $-71.4\%$ \\
          & Steps / episode        & $181.06\,[180.08,182.03]$    & 157.32 & $182.89\,[181.85,183.92]$    & 167.02 & $+1.83$     & $+1.01\%$ & $+9.70$   & $+6.2\%$  \\
          & Falls / episode        & $2.20920\,[2.18351,2.23489]$ & 4.14   & $0.00370\,[0.00325,0.00416]$ & 0.07   & $-2.2055$   & $-99.8\%$ & $-4.07$   & $-98.2\%$ \\
          & Rollbacks / episode    & ---                           & ---    & $3.4385\,[3.3927,3.4843]$     & 7.39   & $+3.4385$   & n/a       & $+7.39$   & n/a       \\
        \midrule
        \multirow{5}{*}{\textbf{Taxi-v3}}
          & Total Reward           & $-1652.93\,[-1656.98,-1648.88]$ & 652.74 & $-567.09\,[-568.75,-565.44]$ & 267.00 & $+1085.84$ & $+65.7\%$ & $-385.74$ & $-59.1\%$ \\
          & Steps / episode        & $681.85\,[680.11,683.60]$        & 281.22 & $698.65\,[696.74,700.56]$    & 308.49 & $+16.80$   & $+2.46\%$ & $+27.27$  & $+9.7\%$  \\
          & Illegal Drops / episode& $110.21690\,[109.95840,110.47540]$ & 41.70 & $0.06940\,[0.06764,0.07116]$ & 0.28   & $-110.1475$ & $-99.9\%$ & $-41.42$  & $-99.3\%$ \\
          & Deliveries / episode   & $0.99410\,[0.99362,0.99458]$      & 0.077  & $0.98500\,[0.98425,0.98575]$ & 0.121  & $-0.00910$  & $-0.92\%$ & $+0.0450$ & $+58.1\%$ \\
          & Rollbacks / episode    & ---                               & ---    & $111.5006\,[111.2280,111.7732]$ & 43.98 & $+111.5006$ & n/a       & $+43.98$  & n/a       \\
        \bottomrule
      \end{tabular}
    \end{adjustbox}%
  }

  \begin{minipage}{\wd0}
    \captionsetup{justification=raggedright,singlelinecheck=off,aboveskip=0pt,belowskip=12pt}
    \caption{Comparison of performance and variance metrics between vanilla \(Q\)-learning and the reversibility-augmented agent.}
    \label{tab:q_mod_table}
    
    \usebox0
  \end{minipage}

\end{adjustwidth}
\end{table}

\subsection{Performance and Safety in \texttt{CliffWalking-v0}}

\begin{enumerate}[label=\arabic*.]
  \item \textbf{Mean Episode Return:}
    The standard \(Q\)-learning agent attains an average return of \(-399.77\) (\(\sigma = 563.78\)), whereas the reversibility-augmented agent achieves \(-179.81\) (\(\sigma = 160.97\)), yielding a \(+55.0\%\) reduction in penalty (\(\Delta = +219.96\)). This indicates that penalizing low-reversibility transitions and undoing unsafe moves steers the policy away from cliff-edge states.

  \item \textbf{Catastrophic Falls:}
    Under vanilla \(Q\)-learning, the agent falls off the cliff \(2.20920\) times per episode (\(\sigma = 4.14\)). Introducing rollbacks reduces falls to \(0.00370\) per episode (\(\sigma = 0.07\))-a \(-99.8\%\) change. The rollback mechanism thus intercepts essentially all cliff transgressions before terminal penalty.

  \item \textbf{Trajectory Efficiency:}
    Despite averaging \(3.4385\) corrective rollbacks per episode (\(\sigma = 7.39\)), the augmented agent’s trajectories change from \(181.06\) steps (\(\sigma = 157.32\)) to \(182.89\) steps (\(\sigma = 167.02\)), a \(+1.01\%\) shift. In this domain, safety comes at negligible path-length cost.

  \item \textbf{Variance Control:}
    Variability in safety-critical quantities contracts sharply: return standard deviation drops by \(71.4\%\) (\(563.78 \rightarrow 160.97\)) and falls variance by \(98.2\%\) (\(4.14 \rightarrow 0.07\)). Path-length variability rises modestly (\(157.32 \rightarrow 167.02\)), consistent with occasional rollback-induced detours while preserving robust safety. This variance reduction effect is consistent with stabilization approaches such as Averaged-DQN \cite{anschel2017averaged}, though our rollback mechanism achieves stability through explicit corrective interventions rather than ensemble averaging.
\end{enumerate}

\subsection{Performance and Safety in \texttt{Taxi-v3}}

\begin{enumerate}[label=\arabic*.]
  \item \textbf{Mean Episode Return:}
    Vanilla \(Q\)-learning yields \(-1652.93\) (\(\sigma = 652.74\)), whereas the rollback-equipped agent reaches \(-567.09\) (\(\sigma = 267.00\)), a \(+65.7\%\) improvement (\(\Delta = +1085.84\)). Preventing illegal transitions before they accrue penalties recovers the bulk of negative reward.

  \item \textbf{Illegal Action Suppression:}
    The frequency of illegal actions plunges from \(110.21690\) per episode (\(\sigma = 41.70\)) to \(0.06940\) (\(\sigma = 0.28\))-a \(-99.9\%\) change. The agent executes an average of \(111.5006\) rollbacks (\(\sigma = 43.98\)) per episode, effectively catching nearly every invalid transition and avoiding the associated \(-10\) penalties and wasted navigation.

  \item \textbf{Trajectory Length and Success Rate:}
    Corrective rollbacks extend trajectories modestly: steps per episode rise from \(681.85\) (\(\sigma = 281.22\)) to \(698.65\) (\(\sigma = 308.49\)), a \(+2.46\%\) increase. Delivery success declines slightly from \(0.99410\) to \(0.98500\) (\(\Delta = -0.00910\), \(-0.92\%\); \(\sigma\): \(0.077 \rightarrow 0.121\)), reflecting a more conservative policy that avoids risky shortcuts.

  \item \textbf{Variance Control:}
    Return variance shrinks by \(59.1\%\) (\(652.74 \rightarrow 267.00\)) and illegal-action standard deviation by \(99.3\%\) (\(41.70 \rightarrow 0.28\)). Step-count variability rises by \(9.7\%\) (\(281.22 \rightarrow 308.49\)), and delivery variability increases (from \(0.077\) to \(0.121\)), attributable to episodic fluctuations in rollback frequency and success outcomes. Overall, safety-critical metrics become markedly more predictable while modestly increasing path-length dispersion. This predictability aligns with prior work on deep exploration methods such as Bootstrapped DQN \cite{osband2016bootstrapped}, though our approach reduces dispersion by constraining unsafe transitions rather than by bootstrapped value-function sampling.
\end{enumerate}

\subsection{Parameter Analysis}

\begin{figure}[H]
  \centering
\includegraphics[width=\linewidth]{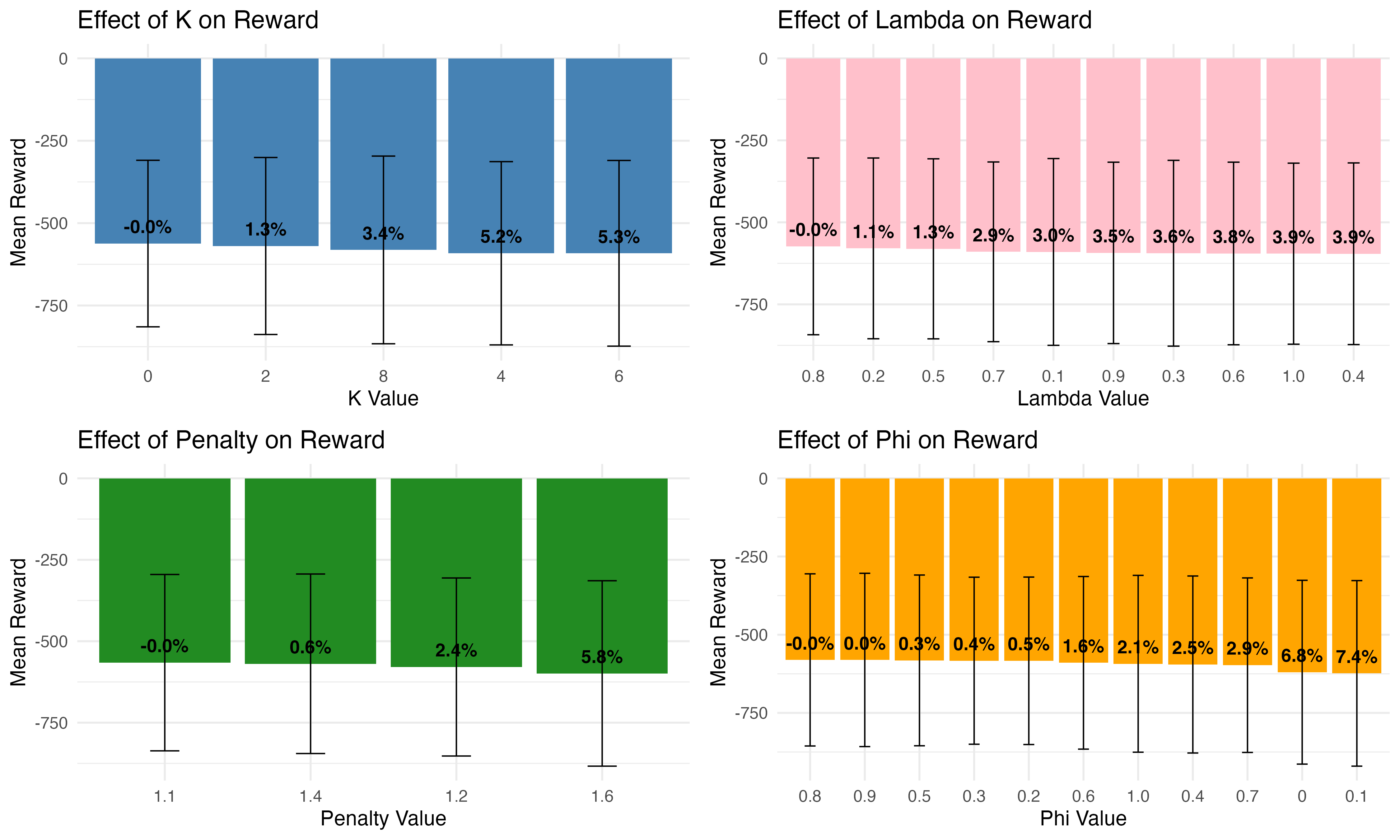} 
  \caption{Parameter sensitivity analysis in \texttt{Taxi-v3}.}
  \label{fig:taxi_parameter}
\end{figure}

\begin{figure}[H]
  \centering
\includegraphics[width=\linewidth]{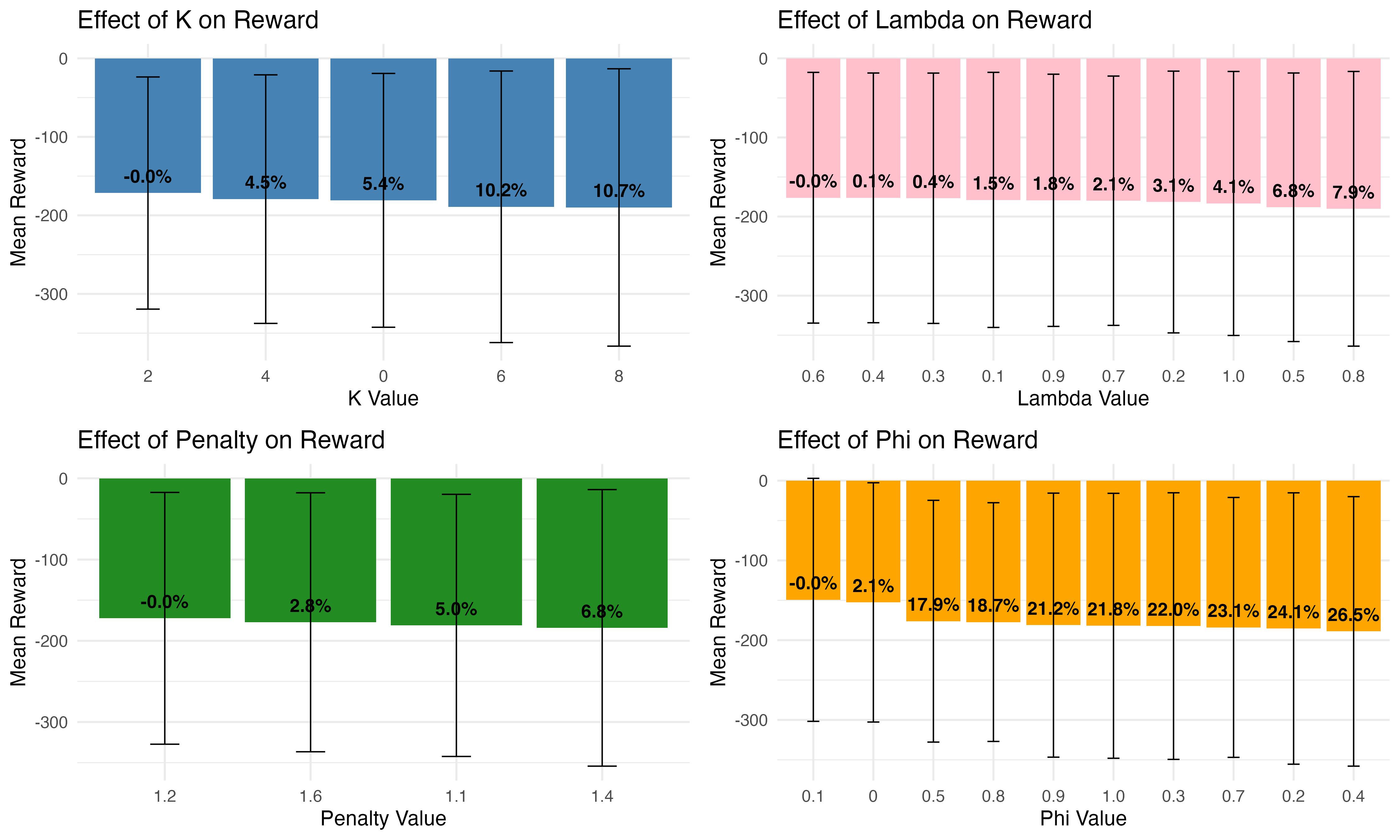}
  \caption{Parameter sensitivity analysis in \texttt{CliffWalking-v0}.}
  \label{fig:cliff_parameter}
\end{figure}

We now examine the sensitivity of the reversible learning framework to its four main parameters: horizon length (\(K\)), precedence learning rate (\(\lambda\)), penalty magnitude, and the initialization value of the reversibility estimator (\(\Phi_0\)). For both domains, the first bar (Fig~\ref{fig:taxi_parameter}, Fig~\ref{fig:cliff_parameter}) in each sweep corresponds to the empirically optimal value, which we interpret before discussing degradation under alternative settings.  

\paragraph{Horizon (\(K\)).}  
In \texttt{CliffWalking-v0}, the optimal value is \(K=2\). This aligns with the environment’s local grid dynamics, where safe reversals are typically only one or two steps away. Shorter windows (e.g., \(K=1\)) miss legitimate reversals and cause excessive rollbacks, while longer horizons (e.g., \(K=4,6\)) dilute the local signal and mistakenly treat cliff-edge detours as reversible. Thus, reversibility in CliffWalking is predominantly local, and \(K=2\) best captures the true return structure.  

By contrast, \texttt{Taxi-v3} exhibits an optimum at \(K=0\). Reversibility here is immediate: illegal pick-ups and drop-offs reveal themselves instantly, and grid navigation is inherently safe. Any extension of the horizon introduces noise from loops in the taxi’s movement, delaying rollback corrections. Performance degrades monotonically with larger \(K\), with \(K=6\)–\(8\) being the weakest. This contrast illustrates that CliffWalking benefits from short local windows, while Taxi rewards purely instantaneous checks.  

\paragraph{Precedence Learning Rate (\(\lambda\)).}  
For \texttt{CliffWalking-v0}, the optimal setting is \(\lambda = 0.6\), with \(0.4\) and \(0.3\) also performing strongly. Smaller values (e.g., \(0.1\)) underfit reversibility signals, while larger extremes destabilize updates. This indicates that CliffWalking favors a relatively fast but stable reversibility learner.  

In \texttt{Taxi-v3}, the optimal value is \(\lambda = 0.8\), with weaker but still viable performance at \(0.2\)–\(0.5\). Too-slow rates again lag behind environmental evidence, while non-optimal values like \(1.0\) or \(0.4\) inject noise. Because Taxi features many repeated sub-trajectories, rapid updates to \(\Phi\) are necessary to keep rollback triggers aligned with the current episode dynamics.  

\paragraph{Penalty Magnitude.}  
In \texttt{CliffWalking-v0}, the best result is at penalty \(= 1.2\), followed closely by \(1.6\) and \(1.1\). The weakest was \(1.4\), with a \(6.8\%\) performance drop relative to the optimum. This suggests that penalties clustered around \(1.1\)–\(1.6\) work well, but tuning is important: too low under-corrects, while poorly calibrated values (like \(1.4\)) disrupt efficiency.  

In \texttt{Taxi-v3}, the optimum is \(1.1\), with \(1.4\) and \(1.2\) still serviceable. However, \(1.6\) produced the weakest performance, over-constraining exploration. Since Taxi already imposes large native penalties (\(-10\) for illegal moves), a lighter reversibility penalty is sufficient; higher values introduce unnecessary rollback frequency.  

\paragraph{Initialization Value (\(\Phi_0\)).}  
The \texttt{CliffWalking-v0} domain is best served by \(\Phi_0 = 0.0\) or \(0.1\). This pessimistic prior reflects the environment’s high asymmetry between safe moves and catastrophic cliff falls. By assuming most transitions are irreversible until proven otherwise, the agent leverages rollback early and avoids premature overconfidence near the cliff. More optimistic initializations (e.g., \(0.5\)–\(1.0\)) performed substantially worse, as they caused misjudgments of danger and frequent falls.  

The opposite holds in \texttt{Taxi-v3}, where the optimal initialization lies around \(\Phi_0 = 0.8\)–\(0.9\). Because Taxi contains many inherently reversible transitions (safe grid navigation), an optimistic prior reduces unnecessary rollbacks and penalties on benign moves. Pessimistic values (e.g., \(\Phi_0 = 0.0\)–\(0.1\)) misclassify ordinary movements as irreversible, inflating rollbacks and hurting efficiency.  

\paragraph{Summary.}  
Taken together, the parameter sweeps reveal environment-specific sensitivities. \texttt{CliffWalking-v0} rewards a short local horizon (\(K=2\)), a moderately fast precedence learner (\(\lambda=0.6\)), a carefully tuned penalty near \(1.2\), and a pessimistic initialization (\(\Phi_0=0.0\)–\(0.1\)) reflecting its hazardous structure. \texttt{Taxi-v3}, in contrast, favors hyper-local checks (\(K=0\)), a fast learner (\(\lambda=0.8\)), a lighter penalty (\(1.1\)), and an optimistic prior (\(\Phi_0=0.8\)–\(0.9\)). These contrasts underscore that reversibility-aware RL is not governed by a single “best” hyperparameter profile but must adapt its biases: CliffWalking demands caution and pessimism near irreversible cliffs, while Taxi thrives with optimism, immediacy, and lighter corrective signals.  

\subsection{Parameter Sensitivity}

The effectiveness of the reversibility + rollback framework critically depends on two key parameters: 

\begin{description}[leftmargin=1em, labelwidth=2.5cm, style=nextline]
  \item[Q-Table Initialization Value (\(Q_0\))]  
    In the modified algorithm, initializing all \(Q\)-values to zero biases the penalty-and-rollback criterion: zero \(Q\)-values can cause the rollback condition to misfire, leading to suboptimal or inconsistent rollbacks. We therefore initialize for both environments
    \[
      Q_0 = -1.
    \]
    This was the optimal initialization value given the reward structure in both domains.

  \item[Penalty Threshold (\(T\))]  
    The rollback criterion fires when the reversibility-penalized TD target falls below 
    \[
      T \cdot Q(s,a).
    \]
    If \(T\) is too high, legitimate exploratory moves are rolled back excessively, over-constraining the policy; if \(T\) is too low, unsafe transitions may slip through uncorrected. We select \(T\) empirically based on the domain’s reward scale (e.g., \(T=3\) for both \texttt{CliffWalking-v0} and \texttt{Taxi-v3}) to balance safety intervention against necessary exploration.

    An incorrect threshold choice can either 
    \begin{enumerate}[label=(\alph*)]
      \item suppress learning by over-rolling back, or
      \item fail to prevent catastrophic events,
    \end{enumerate}
    resulting in skewed performance metrics and increased variance.
\end{description}

\subsection{Ablation Study}
\label{sec:ablation}

We disentangle the effects of three components-\emph{rollback}, \emph{threshold-based penalization}, and \emph{precedence (\(\Phi\)) penalties}-across \texttt{CliffWalking-v0} and \texttt{Taxi-v3}. Agent configurations and hyperparameters are listed in Table~\ref{tab:params_abli}; outcome metrics are reported in Tables~\ref{tab:cliff_results}, \ref{tab:taxi_results}, and the attribution-style summary in Table~\ref{tab:component_contrib}.

\begin{table*}[ht]
\centering
\small
\setlength{\tabcolsep}{4pt}
\renewcommand{\arraystretch}{1.15}

\begin{adjustwidth}{-1.1in}{-0.25in}
  \setbox0=\hbox{%
    \begin{adjustbox}{max width=\linewidth}
      \begin{tabular}{llcccccccccc}
        \toprule
        \textbf{Env} & \textbf{Agent} & $\alpha$ & $\gamma$ & $\epsilon$ & $q\_table\_init$ & $K$ & $\alpha_\phi$ & $\lambda_{prec}$ & $\phi_{init}$ & $threshold$ & $penalty$ \\
        \midrule
        \multirow{8}{*}{\texttt{CliffWalking-v0}} 
         & Baseline (QL)     & 0.1 & 0.99 & 0.1 & $0.0$  & -- & --   & --  & --  & -- & -- \\
         & RollbackOnly      & 0.1 & 0.99 & 0.1 & $-1.0$ & -- & --   & --  & --  & 3  & -- \\
         & ThresholdPeAgent  & 0.1 & 0.99 & 0.1 & $-1.0$ & -- & --   & --  & --  & 3  & 1.1 \\
         & Roll\_Threshold   & 0.1 & 0.99 & 0.1 & $-1.0$ & -- & --   & --  & --  & 3  & 1.1 \\
         & PrecedenceOnly    & 0.1 & 0.99 & 0.1 & $-1.0$ & 2  & 0.01 & 0.6 & 0.1 & -- & -- \\
         & Precedence\_R     & 0.1 & 0.99 & 0.1 & $-1.0$ & 2  & 0.01 & 0.6 & 0.1 & 3  & -- \\
         & Precedence\_Th    & 0.1 & 0.99 & 0.1 & $-1.0$ & 2  & 0.01 & 0.6 & 0.1 & 3  & 1.1 \\
         & FullModel         & 0.1 & 0.99 & 0.1 & $-1.0$ & 2  & 0.01 & 0.6 & 0.1 & 3  & 1.1 \\
        \midrule
        \multirow{8}{*}{\texttt{Taxi-v3}}
         & Baseline (QL)     & 0.1 & 0.99 & 0.1 & $0.0$  & -- & --   & --  & --  & -- & -- \\
         & RollbackOnly      & 0.1 & 0.99 & 0.1 & $-1.0$ & -- & --   & --  & --  & 3  & -- \\
         & ThresholdPeAgent  & 0.1 & 0.99 & 0.1 & $-1.0$ & -- & --   & --  & --  & 3  & 1.1 \\
         & Roll\_Threshold   & 0.1 & 0.99 & 0.1 & $-1.0$ & -- & --   & --  & --  & 3  & 1.1 \\
         & PrecedenceOnly    & 0.1 & 0.99 & 0.1 & $-1.0$ & 2  & 0.01 & 0.8 & 0.8 & -- & -- \\
         & Precedence\_R     & 0.1 & 0.99 & 0.1 & $-1.0$ & 2  & 0.01 & 0.8 & 0.8 & 3  & -- \\
         & Precedence\_Th    & 0.1 & 0.99 & 0.1 & $-1.0$ & 2  & 0.01 & 0.8 & 0.8 & 3  & 1.1 \\
         & FullModel         & 0.1 & 0.99 & 0.1 & $-1.0$ & 2  & 0.01 & 0.8 & 0.8 & 3  & 1.1 \\
        \bottomrule
      \end{tabular}
    \end{adjustbox}%
  }%
  \begin{minipage}{\wd0}
    \captionsetup{justification=raggedright,singlelinecheck=off,aboveskip=0pt,belowskip=5pt}
    \caption{Parameter matrix for agents in \texttt{CliffWalking-v0} and \texttt{Taxi-v3}.}
    \label{tab:params_abli}

    \usebox0
  \end{minipage}
\end{adjustwidth}
\end{table*}

\begin{table}[H]
\centering
\small
\setlength{\tabcolsep}{5pt}
\renewcommand{\arraystretch}{1.15}

\begin{adjustwidth}{-1.1in}{-0.25in}
  \setbox0=\hbox{%
    \begin{adjustbox}{max width=\linewidth}
      \begin{tabular}{lcccccc}
        \toprule
        \textbf{Agent} & \textbf{Reward} & \textbf{$\Delta$ Reward} & \textbf{$\Delta$\%} & \textbf{Failures} & \textbf{$\Delta$ Fail\%} & \textbf{Rollbacks} \\
        \midrule
        Roll\_Threshold  & $-174.4 \pm 151.4$ & $+225.3$ & $+56.4\%$ & $0.004$ & $+99.8\%$ & $2.3$ \\
        RollbackOnly     & $-174.9 \pm 152.3$ & $+224.8$ & $+56.2\%$ & $0.004$ & $+99.8\%$ & $2.4$ \\
        FullModel        & $-179.8 \pm 161.0$ & $+220.0$ & $+55.0\%$ & $0.004$ & $+99.8\%$ & $3.4$ \\
        Precedence\_R    & $-181.5 \pm 162.8$ & $+218.3$ & $+54.6\%$ & $0.004$ & $+99.8\%$ & $3.5$ \\
        ThresholdPeAgent & $-398.2 \pm 566.1$ & $+1.6$   & $+0.4\%$  & $2.174$ & $+1.6\%$  & n/a \\
        Baseline         & $-399.8 \pm 563.8$ & $+0.0$   & $+0.0\%$  & $2.209$ & n/a       & n/a \\
        Precedence\_Th   & $-424.1 \pm 605.4$ & $-24.3$  & $-6.1\%$  & $2.354$ & $-6.6\%$  & n/a \\
        PrecedenceOnly   & $-427.5 \pm 609.3$ & $-27.8$  & $-6.9\%$  & $2.378$ & $-7.7\%$  & n/a \\
        \bottomrule
      \end{tabular}
    \end{adjustbox}%
  }%
 \begin{minipage}{\wd0}
    \captionsetup{justification=raggedright,singlelinecheck=off,aboveskip=5pt,belowskip=5pt}
    \caption{Ablation results on \textsc{CliffWalking\texttt{-}v0}. Rewards are averaged with standard deviation. $\Delta$ values are relative improvements over the baseline.}
    \label{tab:cliff_results}

    \usebox0
  \end{minipage}
\end{adjustwidth}
\end{table}

\begin{table}[H]
\centering
\small
\setlength{\tabcolsep}{5pt}
\renewcommand{\arraystretch}{1.15}

\begin{adjustwidth}{-1.1in}{-0.25in}
  \setbox0=\hbox{%
    \begin{adjustbox}{max width=\linewidth}
      \begin{tabular}{lcccccc}
        \toprule
        \textbf{Agent} & \textbf{Reward} & \textbf{$\Delta$ Reward} & \textbf{$\Delta$\%} & \textbf{Failures} & \textbf{$\Delta$ Fail\%} & \textbf{Rollbacks} \\
        \midrule
        RollbackOnly     & $-551.8 \pm 241.7$ & $+1101.2$ & $+66.6\%$ & $0.033$  & $+100.0\%$ & $110.3$ \\
        Roll\_Threshold  & $-552.0 \pm 241.0$ & $+1101.0$ & $+66.6\%$ & $0.063$  & $+99.9\%$  & $110.2$ \\
        FullModel        & $-567.1 \pm 267.0$ & $+1085.8$ & $+65.7\%$ & $0.069$  & $+99.9\%$  & $111.5$ \\
        Precedence\_R    & $-567.7 \pm 266.0$ & $+1085.2$ & $+65.7\%$ & $0.017$  & $+100.0\%$ & $111.7$ \\
        Baseline         & $-1652.9 \pm 652.7$& $+0.0$    & $+0.0\%$  & $110.217$& n/a        & n/a \\
        ThresholdPeAgent & $-1654.2 \pm 654.1$& $-1.2$    & $-0.1\%$  & $110.269$& $-0.0\%$   & n/a \\
        Precedence\_Th   & $-1683.2 \pm 699.7$& $-30.2$   & $-1.8\%$  & $111.632$& $-1.3\%$   & n/a \\
        PrecedenceOnly   & $-1686.1 \pm 702.1$& $-33.1$   & $-2.0\%$  & $111.805$& $-1.4\%$   & n/a \\
        \bottomrule
      \end{tabular}
    \end{adjustbox}%
  }%
   \begin{minipage}{\wd0}
    \captionsetup{justification=raggedright,singlelinecheck=off,aboveskip=5pt,belowskip=5pt}
    \caption{Ablation results on \textsc{Taxi\texttt{-}v3}. Rewards are averaged with standard deviation. $\Delta$ values are relative improvements over the baseline.}
    \label{tab:taxi_results}

    \usebox0
  \end{minipage}
\end{adjustwidth}
\end{table}

\begin{table}[H]
\centering
\small
\setlength{\tabcolsep}{6pt}
\renewcommand{\arraystretch}{1.15}

\begin{adjustwidth}{-1.1in}{-0.25in}
  \setbox0=\hbox{%
    \begin{adjustbox}{max width=\linewidth}
      \begin{tabular}{llcccc}
        \toprule
        \textbf{Environment} & \textbf{Configuration} & \textbf{Reward Improvement} & \textbf{Share of Full Model} & \textbf{Failure Reduction} & \textbf{Rollbacks / Episode} \\
        \midrule
        \multirow{4}{*}{\textsc{CliffWalking\texttt{-}v0}} 
         & Baseline (Q-Learning)   & $-399.8$ reward, $2.209$ fails & ---      & ---                      & n/a \\
         & Rollback Only           & $+224.8$ ($+56.2\%$)           & $102.2\%$& $+2.205$ ($+99.8\%$)     & $2.4$ \\
         & Precedence Only         & $-27.8$ ($-6.9\%$)             & $-12.6\%$& $-0.169$ ($-7.7\%$)      & n/a \\
         & Full Model (All comps.) & $+220.0$ ($+55.0\%$)           & $100.0\%$& $+2.206$ ($+99.8\%$)     & $3.4$ \\
        \midrule
        \multirow{4}{*}{\textsc{Taxi\texttt{-}v3}} 
         & Baseline (Q-Learning)   & $-1652.9$ reward, $110.217$ fails & ---      & ---                      & n/a \\
         & Rollback Only           & $+1101.2$ ($+66.6\%$)          & $101.4\%$& $+110.184$ ($+100.0\%$)  & $110.3$ \\
         & Precedence Only         & $-33.1$ ($-2.0\%$)             & $-3.1\%$ & $-1.588$ ($-1.4\%$)       & n/a \\
         & Full Model (All comps.) & $+1085.8$ ($+65.7\%$)          & $100.0\%$& $+110.147$ ($+99.9\%$)   & $111.5$ \\
        \bottomrule
      \end{tabular}
    \end{adjustbox}%
  }%
  \begin{minipage}{\wd0}
    \captionsetup{justification=raggedright,singlelinecheck=off,aboveskip=5pt,belowskip=5pt}
    \caption{Component contribution analysis for \textsc{CliffWalking\texttt{-}v0} and \textsc{Taxi\texttt{-}v3}. Baseline refers to vanilla Q-learning without rollback, threshold, or $\Phi$-penalty.}
    \label{tab:component_contrib}

    \usebox0
  \end{minipage}
  
\end{adjustwidth}
\end{table}

We ablate three components-explicit rollback, threshold-based scaling, and precedence ($\Phi$) penalties-across \texttt{CliffWalking-v0} and \texttt{Taxi-v3} under identical tabular Q-learning settings and training budgets. Metrics are mean return, failure rate (falls or illegal actions), rollback frequency, and dispersion (SD), computed over $100{,}000$ episodes per agent.

Rollback is the dominant driver of both safety and performance: \textsc{RollbackOnly} and \textsc{Roll\_Threshold} recover essentially all of the full model’s reward improvement while virtually eliminating failures ($\ge\!99.8\%$). By contrast, \textsc{PrecedenceOnly} underperforms vanilla Q-learning in both domains, indicating that $\Phi$-penalties alone misguide updates when self-transitions and resets are frequent. Thresholding is secondary: it contributes little on its own and adds value primarily when paired with rollback, with gains that depend on the environment.

In \texttt{CliffWalking-v0}, \textsc{Roll\_Threshold} achieves the best mean return ($-174.4$) with \textsc{RollbackOnly} a close second ($-174.9$). Both exceed \textsc{FullModel} ($-179.8$) while maintaining the same near-zero failure rate. Notably, \textsc{Roll\_Threshold} attains the top return with fewer rollbacks per episode (2.3) than \textsc{FullModel} (3.4), suggesting that once catastrophic moves are suppressed, the threshold prunes unnecessary reversions and slightly improves path efficiency.

In \texttt{Taxi-v3}, \textsc{RollbackOnly} is strongest ($-551.8$), narrowly ahead of \textsc{Roll\_Threshold} and clearly ahead of \textsc{FullModel}. Adding $\Phi$ reduces returns without measurable safety gains (failures are already $\approx 0$ under rollback). Thresholding does not meaningfully change rollback usage (110.2 vs 110.3), indicating limited leverage in navigation-dominated regimes where frequent, benign reversions are intrinsic to task structure.

Across both tasks, rollback variants markedly compress reward dispersion (e.g., Cliff SD $\approx\!151$–162 vs Baseline $\approx\!564$; Taxi SD $\approx\!241$–267 vs Baseline $\approx\!653$), consistent with smoother learning trajectories. This variance collapse, coupled with order-of-magnitude failure reductions, supports the view that reversible corrections prevent catastrophic updates from propagating.

Plotting return against rollbacks per episode yields a Pareto-like frontier dominated by rollback agents. In \texttt{CliffWalking-v0}, \textsc{Roll\_Threshold} occupies a favorable corner (better return and fewer rollbacks than \textsc{FullModel}). In \texttt{Taxi-v3}, the frontier is essentially flat between \textsc{RollbackOnly} and \textsc{Roll\_Threshold}, implying that thresholding adds little efficiency once rollback usage saturates.

Mechanistically, rollback acts as a local safety filter that caps downside by immediately reversing low-quality transitions before value errors spread-akin to a risk-sensitive control at the transition level. Thresholding regulates the rollback budget, helping in cliff-like domains where failures are sparse but costly. $\Phi$-penalties pressure the agent away from high-precedence (hard-to-undo) regions, but in environments with many benign self-transitions (e.g., \texttt{Taxi-v3}) this shaping conflates necessary loops with hazards, degrading policy quality unless guarded by rollback.

Practical guidance: use explicit rollback as the default safety primitive; add thresholding to trim extraneous reversions in cliff-like tasks; apply $\Phi$-penalties sparingly and only alongside rollback, tuning them with awareness of self-transition prevalence. This recipe preserves the safety guarantee, captures most of the performance lift, and controls variance.

\section{Discussion}

The results show that embedding reversibility into reinforcement learning improves both safety and performance across environments. In \texttt{CliffWalking-v0}, reversibility-aware agents reduced catastrophic failures by \(\geq 99.8\%\) while substantially improving cumulative returns and compressing variance. In \texttt{Taxi-v3}, selective rollback suppressed illegal actions by \(\geq 99.9\%\), transforming persistent penalties into recoverable states. Together, these outcomes indicate that reversibility not only mitigates overestimation-induced errors but also acts as a variance-control mechanism that stabilizes learning in safety-critical domains.

Ablations isolate \emph{rollback} as the primary driver of these gains. Rollback-only and rollback+threshold agents recover essentially all of the full model’s return improvements while preserving the near-elimination of failures. By contrast, \emph{precedence} (\(\Phi\)) penalties without rollback underperform vanilla \(Q\)-learning in both tasks; with rollback, their contribution is \emph{environment-dependent}: in CliffWalking they are at best marginally helpful (and sometimes neutral) once failures are already suppressed, whereas in Taxi they tend to degrade returns due to frequent benign self-transitions being misclassified as undesirable. Hence, hard interventions (explicit undo) dominate soft shaping; and any \(\Phi\)-based shaping should be used sparingly and only alongside rollback.

Parameter sensitivity reinforces this environment-specific picture. CliffWalking benefits from pessimistic priors on reversibility, short horizons, and moderate penalties-consistent with highly asymmetric costs (safe moves vs.\ cliff falls). Taxi favors optimistic priors, immediate rollbacks, and lighter penalties, reflecting its abundance of inherently reversible transitions and the large native penalty already attached to invalid actions. These contrasts confirm that reversibility-aware RL is not “one-size-fits-all”: it should encode environment-specific biases to trade off caution and efficiency.

Our results are limited to tabular Gym/Gymnasium toy-text domains (e.g., \texttt{CliffWalking-v0}, \texttt{Taxi-v3}), selected for transparency and controllability rather than representational richness. Consequently, conclusions about safety and variance reduction may not carry over without modification to function-approximation settings or high-dimensional continuous control. The rollback operator further assumes access to a safe \emph{previous-state} primitive (or an equivalent reset/checkpoint facility). While this is realistic in simulated grids, it can be non-trivial in real systems; even when available. Finally, effectiveness is sensitive to environment-aware hyperparameter selection---horizon \(K\), threshold \(T\), penalty scale \(\lambda\), and \(\Phi\) initialization (\(\Phi_0\)) which requires tuning prior to deployment or extension to deep function approximation.

Future research should focus on experimenting with the integration of Rollback in function approximation settings and expanding the experimental domains for precedence estimation to narrow down the use cases for precedence estimation usability in terms of performance. Furthermore, this work can be considered a foundation for behavior modeling, as precedence and rollback can be utilized in encoding agent behavior profiles as optimistic, pessimistic, high or low tolerance to risk, and patience level modeling, which provides foundations for conditions in the human decision-making process. 

\section{Conclusion}

We introduce a reversible reinforcement learning framework that couples an empirical reversibility estimator with an explicit rollback operator and, across two benchmark environments, delivers (1) substantial safety gains—over 99\% fewer catastrophic failures and illegal actions, (2) improved performance—roughly 55--66\% higher cumulative reward than vanilla \(Q\)-learning, (3) variance control—markedly lower dispersion in both reward and safety metrics, and (4) environment-specific adaptability—distinct optimal parameterizations for hazardous versus benign domains. Ablations identify rollback as the critical mechanism; thresholding further improves rollback efficiency in cliff-like tasks, while precedence estimation is supportive, strongly context dependent, and harmful if applied without rollback. Overall, reversibility emerges as a practical, powerful organizing principle for safety-sensitive RL. Future work should extend the framework to deep function approximation, develop adaptive hyperparameter tuning across environments, and investigate real-world analogues of rollback in robotics and decision-support systems. By operationalizing the ability to “undo” mistakes, reversibility-aware RL advances the design of safe, robust, and trustworthy autonomous agents and also can be see as foundation for behavior modeling in decision making agents.

\bibliography{references}

\end{document}